\documentclass{article}
\usepackage{spconf,amsmath,epsfig}
\usepackage{enumitem}
\usepackage{amsfonts,amssymb}
\usepackage{bbding}
\usepackage{threeparttable}

\let\OLDthebibliography\thebibliography
\renewcommand\thebibliography[1]{
  \OLDthebibliography{#1}
  \setlength{\parskip}{0pt}
  \setlength{\itemsep}{0pt plus 0.3ex}
}

\begin{document}\sloppy

\def\x{{\mathbf x}}
\def\L{{\cal L}}

\title{multi-level memory-augmented appearance-motion correspondence framework for video anomaly detection}
%
\name{Xiangyu Huang$^{1}$, Caidan Zhao$^{1}$, Jinghui Yu$^{1}$, Chenxing Gao$^{1}$, and Zhiqiang Wu$^{2}$}

\address{$^{1}$ School of Informatics, Xiamen University \\
$^{2}$ PKU-Wuhan Institute for Artificial Intelligence}

\maketitle

\begin{abstract}
Frame prediction based on AutoEncoder plays a significant role in unsupervised video anomaly detection. Ideally, the models trained on the normal data could generate larger prediction errors of anomalies. However, the correlation between appearance and motion information is underutilized, which makes the models lack an understanding of normal patterns. Moreover, the models do not work well due to the uncontrollable generalizability of deep AutoEncoder. To tackle these problems, we propose a multi-level memory-augmented appearance-motion correspondence framework. The latent correspondence between appearance and motion is explored via appearance-motion semantics alignment and semantics replacement training. Besides, we also introduce a Memory-Guided Suppression Module, which utilizes the difference from normal prototype features to suppress the reconstruction capacity caused by skip-connection, achieving the tradeoff between the good reconstruction of normal data and the poor reconstruction of abnormal data. Experimental results show that our framework outperforms the state-of-the-art methods, achieving AUCs of 99.6\%, 93.8\%, and 76.3\% on UCSD Ped2, CUHK Avenue, and ShanghaiTech datasets.
\end{abstract}
\begin{keywords}
Video anomaly detection, unsupervised learning, AutoEncoder, memory network
\end{keywords}
\section{Introduction}
\label{sec:intro}

Video anomaly detection (VAD) aims to identify behaviors or appearance patterns that do not conform to expectations in surveillance videos \cite{chandola2009anomaly}. With the popularity of surveillance equipment in recent years, VAD, a technology that can interpret surveillance video content automatically, has attracted growing interest from academic and industrial societies. VAD has been researched for decades but remains an extremely challenging task because anomalies are difficult to collect for training and the form of anomaly is inherently ambiguous. Therefore, most existing VAD methods are unsupervised \cite{liu2018future, gong2019memorizing, park2020learning, cai2021appearance}, which train an unsupervised learning model can well describe normal events. Events that deviate from the normality model are deemed as anomalies.

Prediction-based methods are a prevalent VAD paradigm by making full use of temporal characteristics of video frames. Based on the assumption that anomalies are unexpected and the model learned on only normal events cannot fit anomalies that have not been seen before \cite{liu2018future}, prediction-based methods train autoencoders (AEs) on normal data to generate accurate predictions (i.e., future frames) for normal events and utilize prediction errors for anomaly measuring. However, this assumption does not always hold true. \\
\indent
On the one hand, existing methods are highly dependent on the local context information of frame sequences and lack an understanding of normality. Nowadays, some researchers explore the semantic attributes related to abnormal events to further improve the performance of VAD. Video anomalies typically consist of abnormal appearance and irregular motion patterns. Two-stream structures \cite{cai2021appearance, nguyen2019anomaly, yu2020cloze, liu2021hybrid} are proposed to separately model appearance and motion patterns and become a mainstream architecture for VAD. For example, Nguyen et al. \cite{nguyen2019anomaly} propose a model consisting of a shared encoder and two separate decoders for frame reconstruction and optical flow prediction tasks separately. Liu et al. \cite{liu2021hybrid} seamlessly combine optical flow reconstruction and frame prediction tasks so that the deviation of modeling irregular motion information further affects the generation of appearance patterns. Then the deviations of two normal patterns can be used jointly to calculate a more accurate anomaly score. These methods can detect anomalies well in most cases but still lack the exploration of the correlation between appearance and motion information. Some anomalies must be detected by considering the correlation between appearance and motion. For instance, on the Avenue dataset \cite{lu2013abnormal}, it is normal for a person walks with a bag, but it is abnormal to throw the bag. From the perspective of appearance alone, people and the bag are both regular objects. Furthermore, people walking on the pavement and the movement of the bag are regular cases from the motion alone. So ignoring the correlation makes the anomaly detector fail on these anomalies. \\
\indent
On the other hand, existing methods rely on deep neural networks with strong representation capacity to model the diverse patterns but are prone to the curse of `overgeneralizing', where abnormal video frames can also be predicted well, indicating that the prediction errors may not be discriminative enough to detect the anomalies \cite{gong2019memorizing, park2020learning, munawar2017limiting}. Specifically, due to the loss of detailed information during down-sampling in AEs, the output image is often blurry. The skip-connection \cite{skipconnect} is always implemented to make the model generate small prediction errors for normal data in the VAD task. However, it also results in a good generation of anomalies. To enlarge the gap of prediction errors between the normal and abnormal samples, some memory-based approaches \cite{gong2019memorizing, park2020learning, cai2021appearance, liu2021hybrid, liu2022learning} have been proposed. They utilize an external memory bank to store the normality during training, then use the memorized normality to boost the prediction of normal data while suppressing the abnormal ones. However, the gap of prediction error can not be effectively enlarged in the existing memory-augmented works \cite{hou2021divide, lv2021learning}. We argue that the weakness of existing memory-based approaches lies in the fact that they regenerate the feature maps in a per-pixel manner. This way is extremely memory-consuming for storing the normal prototypes as memory items across the whole training set. And a small-sized memory may limit the reconstruction capability for normal data, resulting in poor performance for VAD. \\
\indent
To alleviate the above limitations, we propose a multi-level memory-augmented appearance-motion correspondence framework for video anomaly detection, which is illustrated in Figure 1. Specifically, the two-stream encoder takes both video frames and corresponding optical flows as the input and is trained to extract the appearance and motion features separately. Based on the explicit correspondence that appearance and motion signals have common behavioral semantics, we propose a novel appearance-motion semantics alignment loss and use the motion feature instead of the appearance feature to predict future frames for the purpose of modeling the consistent correlation. Moreover, in order to alleviate the `overgeneralizing' in AEs, we introduce a multi-scale Memory-Guided Suppression Module (MGSM) and implement it into skip-connection, which is the major
insecurity for controlling the reconstruction capacity. Unlike previous memory-based works, we propose to utilize the memorized normality to suppress the representation of the encoded features instead of regenerating them. MGSM uses multi-scale features of different encoding layers as queries to retrieve the most relevant items in the corresponding memory banks. Then the similarity to those items, as a suppressor, is multiplied by the corresponding encoded features, which are sent to the decoder through the corresponding skip-connection. This way, avoid relying on a large number of memory items to precisely model the normality. We summarize our contributions as follows: 

\begin{itemize}[leftmargin=*]
  \item We propose a multi-level memory-augmented appearance-motion correspondence framework that uses the appearance and motion semantics consistent correlation gap between normal and abnormal data to spot anomalies. 
  \item We introduce a multi-scale Memory-Guided Suppression Module, which achieves the tradeoff between the good reconstruction on normal data and the poor reconstruction of abnormal data.
  \item Extensive experiments on three benchmark datasets demonstrate the proposed framework outperforms the state-of-the-art methods.
\end{itemize}

\section{Methodology}
As shown in Figure1 (a), the proposed multi-level memory-augmented appearance-motion correspondence framework consists of three parts: A two-stream encoder, a decoder, and a multi-scale Memory-Guided Suppression Module (MGSM). We take video frame clips and corresponding optical flows into the two-stream encoder. The appearance autoencoder $E_{\varphi}$ is trained to learn appearance features, and the motion autoencoder $E_{\theta}$ is to learn motion features. The proposed appearance-motion semantics alignment loss acts on the appearance and motion features of the bottleneck encoding layer and we feed the motion feature instead of the appearance feature into the decoder $D_{\delta}$ to predict the future frame, where the relation between appearance and motion information is established. Meanwhile, multi-scale down-sampling features of multiple encoding layers are leveraged by the proposed MGSM to strengthen the prediction of normal data while suppressing the abnormal ones. All the components are presented in the following subsections in detail. For more details about the architecture, please see the \textbf{appendix}.

\subsection{Appearance and Motion Correlation Modeling}
Given $T$ consecutive frames $\lbrace$$I_{1}$,$I_{2}$,...,$I_{t-1}$,$I_{t}$$\rbrace$, the predicted frame $\hat{I}_{t+1}$ is generated by inputting the first $T$ frames. The corresponding optical flows $\lbrace$$F_{1}$,$F_{2}$,...,$F_{t-1}$,$F_{t}$$\rbrace$ are got by a pre-trained FlowNet \cite{ilg2017flownet}, which have the motion information of foreground objects while ignoring background in previous $t$ frames. The encoder $E_{\varphi}$ compresses the input RGB frames $I_{1:t}$ into the appearance features $f_{a}$ and the encoder $E_{\theta}$ obtains the motion features $f_{m}$, where the appearance and motion patterns are modeled separately. Considering the sparsity of $f_{a}$ and $f_{m}$ due to the use of ReLU activation function, we minimize the cosine distance between $f_{a}$ and $f_{m}$ to explicitly align the semantics of appearance and motion features, as follows:
\begin{equation}
  \underset{E_{\varphi},E_{\theta}}{\min} \left(1 - cosine(f_{a}, f_{m})\right)
\end{equation}
In addition, for the purpose of modeling the explicit relation that appearance and motion signals have common behavioral semantics, we replace the appearance features $f_{a}$ with the motion features $f_{m}$ into the decoder $D_{\delta}$ to predict the future frame $\hat{I}_{t+1}$. Owing to only the motion information of foreground objects in $f_{m}$, it would alleviate the side effects of the complex background and also directly establish the complementary relation between appearance and motion.

\begin{figure*}[!htbp]
\centering
\includegraphics[width=160mm]{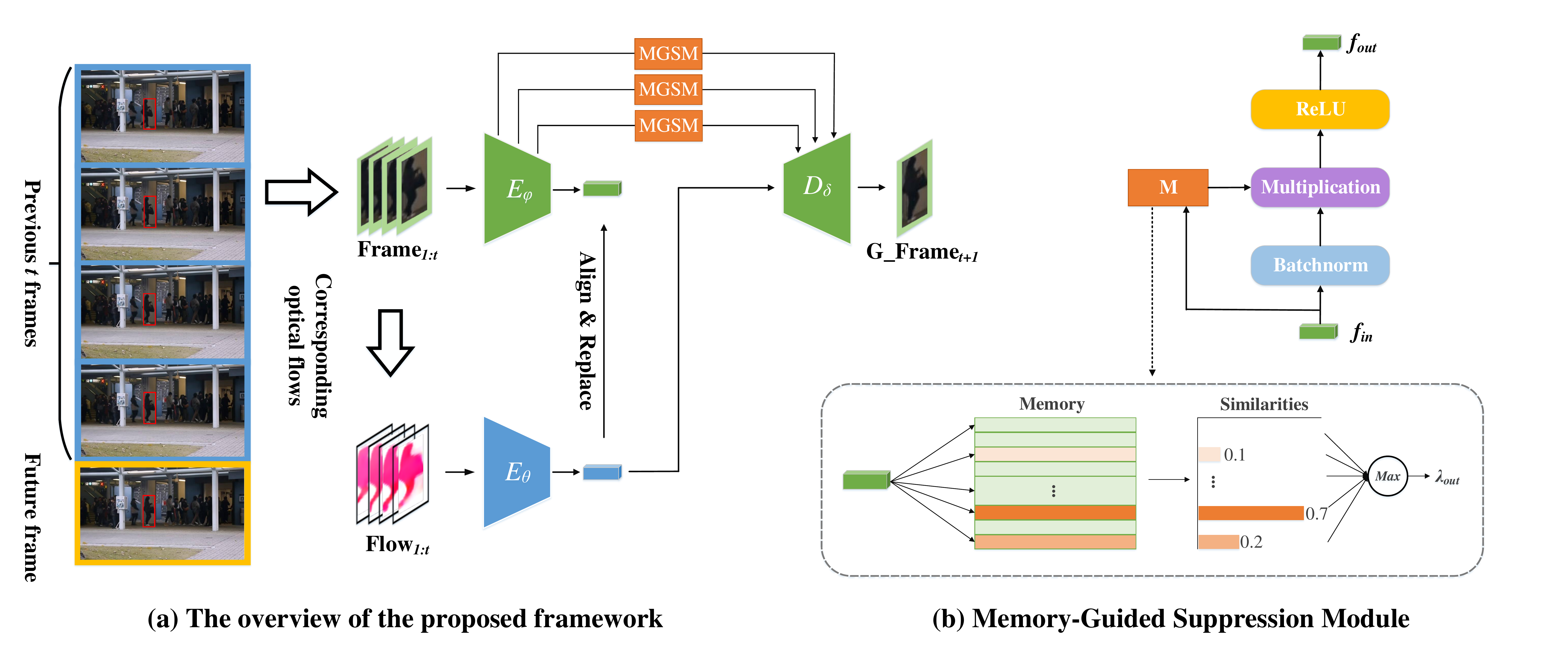}
\caption{Overview of the proposed multi-level memory-augmented appearance-motion correspondence framework.
}
\label{fig:overview}
\end{figure*}

\subsection{Memory-Guided Suppression Module}
The architecture of the proposed MGSM is illustrated in Figure 1 (b). MGSM is trained to learn and compress multiple prototypes that store the normality across the training set. We equip all skip-connections between the encoder $E_{\varphi}$ and the decoder $D_{\delta}$ with the MGSM. MGSM uses the similarity between the input encoded feature and memorized normal prototypes to calculate a suppressor, which suppresses the representations of the corresponding encoded feature. It avoids learning an identity mapping from input to output, which would make normality and anomaly inseparable.

Specifically, given the input consecutive frames, the encoder $E_{\varphi}$ emits a set of feature map $\lbrace$$f_{1}$,$f_{2}$,...,$f_{L-1}$$\rbrace$, where $L$ is the number of down-sampling scales. For each feature map $f_{i}$ of size $H^{i}$ × $W^{i}$ × $C^{i}$, we flatten it into a vector and serve as a query $\mathbf{q}^{k}$ to the memory bank. For the memory bank, we follow \cite{gong2019memorizing} to implement it. Each memory bank is a learnable matrix $\mathbf{M} \in \mathbb{R}^{N \times D}$ consisting of $N$ memory items of fixed dimension $D$ to record the prototypical normal patterns during training. The $j$-th row vector of $\mathbf{M}$ is one memory item $\mathbf{m}_{j} \in \mathbb{R}^{D}$ ($j \in \lbrace 1, 2,..., N \rbrace$). Similar to \cite{gong2019memorizing, liu2021hybrid}, memorized normality is addressed by calculating the attention weight $\mathbf{w}$ based on the similarity between the query $\mathbf{q}^{k}$ and each item $\mathbf{m}_{j}$, as follows:
\begin{equation}
w_j=\frac{\exp \left(\frac{\mathbf{q}^k \mathbf{m}_j^{\top}}{\left\|\mathbf{q}^k\right\|\left\|\mathbf{m}_j\right\|}\right)}{\sum_{v=1}^N \exp \left(\frac{\mathbf{q}^k \mathbf{m}_v^{\top}}{\left\|\mathbf{q}^k\right\|\left\|\mathbf{m}_v\right\|}\right)}
\end{equation}
Based on the attention weights $\mathbf{w} \in \mathbb{R}^{1 \times N}$, a suppressor $\lambda$ is get. Instead of linear combination over the memory items in previous works, the new feature $\hat{\mathbf{q}}^{k}$ that is directly used for the decoder by skip-connection is recomputed by
\begin{equation}
\begin{aligned}
\hat{\mathbf{q}}^k &=\sigma \left(\lambda norm(\mathbf{q}^k)\right) \\
&=\sigma \left(max(\mathbf{w}) norm(\mathbf{q}^k)\right)
\end{aligned}
\end{equation}
where $\sigma$ denotes ReLU activation function, $max$ returns the maximum value in the input attention weights $\mathbf{w}$, and $norm$ denotes the batch normalization. MGSM achieves controllable reconstruction capacity by suppressing the representation of the encoded features instead of regenerating them, alleviating unstable VAD performance due to high dependence on the memory size.

\subsection{Training Loss}
Following previous VAD works based on future frame prediction \cite{liu2018future,liu2021hybrid, liu2022learning}, we use intensity and gradient difference to make the prediction close to its ground truth. The intensity loss guarantees the similarity of pixels between the prediction and its ground truth, and the gradient loss can sharpen the predicted images. Specifically, we minimize the $\ell_{2}$ distance between the predicted frame $\hat{I}$ and its ground truth $I$ as follows:
\begin{equation}
	L_{i n t}=\left\|\hat{I}-I\right\|_{2}^{2}
\end{equation}
The gradient loss is defined as follows:
\begin{equation}
	\begin{aligned}
		L_{g d}=\sum_{i, j}&\left\|\left|\hat{I}_{i, j}-\hat{I}_{i-1, j}\right|-\left|I_{i, j}-I_{i-1, j}\right|\right\|_{1}\\
		+&\left\|\left|\hat{I}_{i, j}-\hat{I}_{i, j-1}\right|-\left|I_{i, j}-I_{i, j-1}\right|\right\|_{1}
	\end{aligned}
\end{equation}
where $i$, $j$ denote the spatial index of a video frame.
The proposed appearance-motion semantics alignment loss is defined as follows:
\begin{equation}
	L_{align}=1-\frac{\langle{f_{a}}, f_{m}\rangle}{\|f_{a}\left\|_{2}\right\|f_{m}\|_{2}}
\end{equation}
where$f_{a}$, $f_{m}$ are the appearance and motion features of the bottleneck encoding layer, respectively.
Additionally, we also design two feature loss to make the learned normal prototypes have the properties of compactness and diversity. The compactness loss and diversity loss are defined as follows:
\begin{equation}
\mathcal{L}_{comp}=\sum_{i=1}^N\left\|\boldsymbol{q}^i-\boldsymbol{m}_1^i\right\|_2^2
\end{equation}

\begin{equation}
\mathcal{L}_{diver}=\sum_{i=1}^N\left\|\boldsymbol{q}^i-\boldsymbol{m}_1^i\right\|_2^2-\left\|\boldsymbol{q}^i-\boldsymbol{m}_2^i\right\|_2^2
\end{equation}
where $\boldsymbol{m}_1^i$ and $\boldsymbol{m}_2^i$ denote the first and second nearest memory items to query $\boldsymbol{q}^i$.
Finally, the overall loss $L$ for training takes the form as follows:
\begin{equation}
    \begin{aligned}
	L&=\lambda_{int} L_{i n t}+\lambda_{g d} L_{g d}+ \lambda_{align} L_{align} \\
	&+ \lambda_{comp} L_{comp} + \lambda_{diver} L_{diver} + \lambda_{model} \left\|W\right\|_{2}^{2}
    \end{aligned}
\end{equation}
where $\lambda_{int}$, $\lambda_{gd}$, $\lambda_{align}$, $\lambda_{comp}$, and $\lambda_{diver}$ are balancing hyper-parameters, $W$ is the parameter of the model, and $\lambda_{model}$ is a regularization hyper-parameter that controls the model complexity. 

\subsection{Anomaly Detection}
Our anomaly score is composed of two parts during the testing phase: (1) the future frame prediction error $S_{p}=\left\|\hat{I}-I\right\|_{2}^{2}$ and (2) the inconsistency of appearance and motion feature $S_{f}=1-\frac{\langle{f_{a}}, f_{m}\rangle}{\|f_{a}\left\|_{2}\right\|f_{m}\|_{2}}$. Then, we get the final anomaly score by fusing the two parts using a weighted sum strategy as follows:
\begin{equation} \label{e5}
	\mathrm{S}=w_{p} \frac{S_{p}-u_{p}}{\delta_{p}}+w_{f} \frac{S_{f}-u_{f}}{\delta_{f}}
\end{equation}
where $u_{p}$, $\delta_{p}$, $u_{f}$, and $\delta_{f}$ denote the means and standard deviations of prediction error and the inconsistency between appearance and motion feature of all the normal training samples. $w_{p}$ and $w_{f}$ represent the weights of the two scores.

\section{Experiments and Results}

 \subsection{Datasets and Evaluation Criterion}

To evaluate the performance of the proposed framework, we conduct experiments on three benchmark datasets, \emph{i.e.}, UCSD Ped2 \cite{li2013anomaly}, CUHK Avenue \cite{lu2013abnormal} and ShanghaiTech \cite{luo2017revisit}.

Following most previous VAD works \cite{liu2018future, cai2021appearance, liu2021hybrid}, we employ frame-level area under the curve (AUC) of the receiver operation characteristic as an evaluation metric. The receiver operation characteristic curve is measured by varying the threshold over the anomaly score. Higher AUC values represent better performance for anomaly detection. 

\begin{figure*}[!htbp]
\centering
\includegraphics[width=150mm]{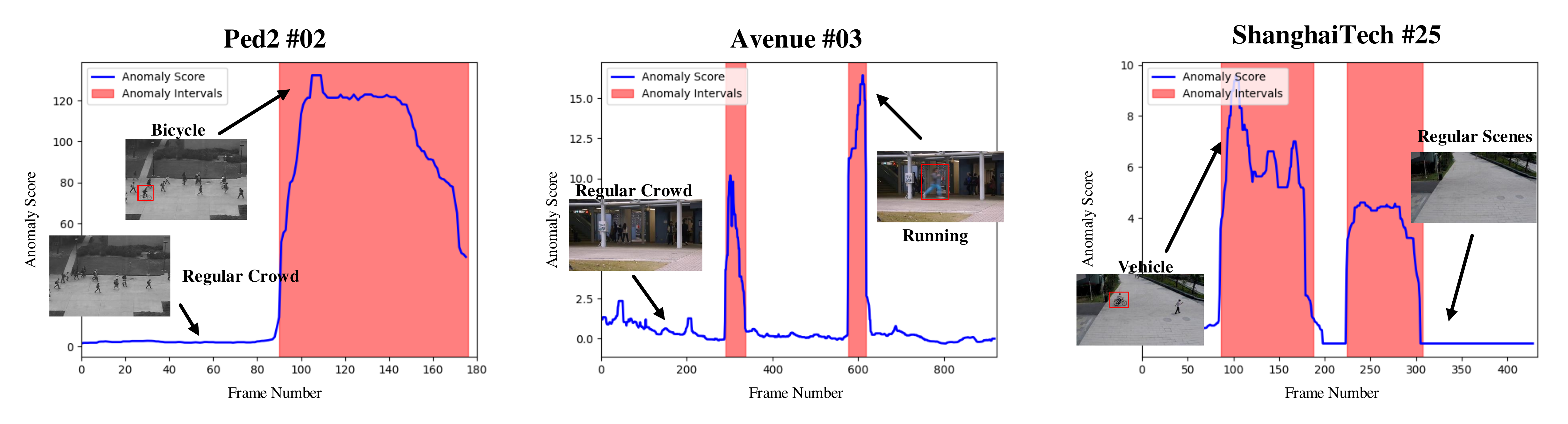}
\caption{Anomaly score curves of some testing video clips.
Red regions represent ground truth anomalous frames.
}
\label{fig:curve}
\end{figure*}

\subsection{Parameters and Implementation Details}
Similar to \cite{yu2020cloze, liu2021hybrid, zhou2022object}, we utilize the patches with foreground objects to train our framework. Specifically, all foreground objects are extracted from original videos. RoI bounding boxes identify foreground objects. For each RoI, a spatial-temporal cube (STC) \cite{yu2020cloze} composed of the object in the current frame and the content in the same region of previous 4 frames will be built. The STCs are resized to 32 × 32 pixels. The corresponding optical flows are generated by FlowNet2 \cite{ilg2017flownet}, and the STCs for optical flows are obtained in a similar way. There are many objects in a frame, so we select the maximum anomaly score of all objects as the final anomaly score.

The proposed framework is trained using the Pytorch framework with an NVIDIA RTX 3090 GPU. Adam is used as the optimizer, and the initial learning rate is set to $2e^{-4}$, decayed by 0.8 after every ten epochs. The batch size and epoch number of Ped2, Avenue, and ShanghaiTech are set to $(128, 60)$, $(128, 30)$, $(384, 30)$, respectively. The multi-scale memory sizes of MGSM for Ped2, Avenue and ShanghaiTech are $(40, 40, 40)$, $(40, 40, 40)$, and $(40, 60, 80)$, respectively. The hyper-parameters $\lambda_{int}$, $\lambda_{gd}$, $\lambda_{align}$, $\lambda_{comp}$, $\lambda_{diver}$ and $\lambda_{model}$ for Ped2, Avenue, and ShanghaiTech are set to $(1, 1, 1, 5e^{-3}, 1e^{-4}, 1)$, $(1, 1, 1, 5e^{-3}, 1e^{-4}, 1)$, $(1, 1, 10, 5e^{-3}, 1e^{-4}, 1)$, respectively. Then the error fusing weights $(w_{p}, w_{f})$ for Ped2, Avenue, and ShanghaiTech are set to $(0.2, 0.8)$, $(0.7, 0.3), (0.3, 0.7)$, respectively.

\begin{table}
\footnotesize
    \begin{center}
	\caption{AUROC (\%) comparison between the proposed method and state-of-the-art VAD methods on three benchmark datasets.}
	\label{t1}
	\begin{tabular}{c|ccc}
    \hline
    Methods            &   UCSD Ped2 & CUHK Avenue & ShanghaiTech \\ \hline
			 ConvLSTM-AE\cite{luo2017remembering} & 88.1          & 77            & N/A           \\
                Frame-Pred.\cite{liu2018future}  & 95.4          & 85.1          & 72.8          \\
			 MemAE\cite{gong2019memorizing}       & 94.1          & 83.3          & 71.2          \\
                AMC\cite{nguyen2019anomaly}     & 96.2          & 86.9             & N/A         \\
			 MNAD-R\cite{park2020learning}      & 97            & 88.5          & 70.5          \\
			 VEC\cite{yu2020cloze}         & 97.3          & 90.2          & 74.8          \\
			 MPU\cite{lv2021learning}         & 96.9          & 89.5          & 73.8           \\
			 AMMC-Net\cite{cai2021appearance}    & 96.6          & 86.6          & 73.7          \\
                STM-AE\cite{liu2022learning}      & 98.1         & 89.8          & 73.8        \\
			 OGMR-Net\cite{zhou2022object}     & 97.4          & 92.6          & 74.9          \\ \hline
    \textbf{Our method}            & \textbf{99.6}      & \textbf{93.8}       & \textbf{76.3}        \\ \hline
	\end{tabular}
	\end{center}
\end{table}

\subsection{Comparison with State-of-the-art Methods}
To our best knowledge, we compare the proposed framework with the state-of-the-art (SOTA) VAD methods on the UCSD ped2 \cite{li2013anomaly}, CUHK Avenue \cite{lu2013abnormal} and ShanghaiTech \cite{luo2017revisit} datasets. The results are summarized in Table \ref{t1}. Our framework outperforms compared SOTA methods on all three benchmarks. Compared with two-stream based methods (\emph{e.g.} AMC\cite{nguyen2019anomaly}, VEC\cite{yu2020cloze}, and AMMC-Net\cite{cai2021appearance}), our method achieves notable improvement, indicating that our exploring correlation between appearance and motion is more effective for unsupervised VAD. In particular, compared to memory-based methods (\emph{e.g.} MemAE \cite{gong2019memorizing}, MNAD-R\cite{park2020learning}, MPU\cite{lv2021learning}, and STM-AE\cite{liu2022learning}), our method gets a significant improvement on the challenging ShanghaiTech datasets, demonstrating that we alleviate the limitations of using the memory bank for complex scenes.

The qualitative results are shown in Figure 2 and Figure 3, including anomaly curves of some testing videos and some visualization examples on three benchmark datasets, which intuitively demonstrate the effectiveness of our method.

\begin{figure}[ht]
	\centering
	\includegraphics[width=6.5cm]{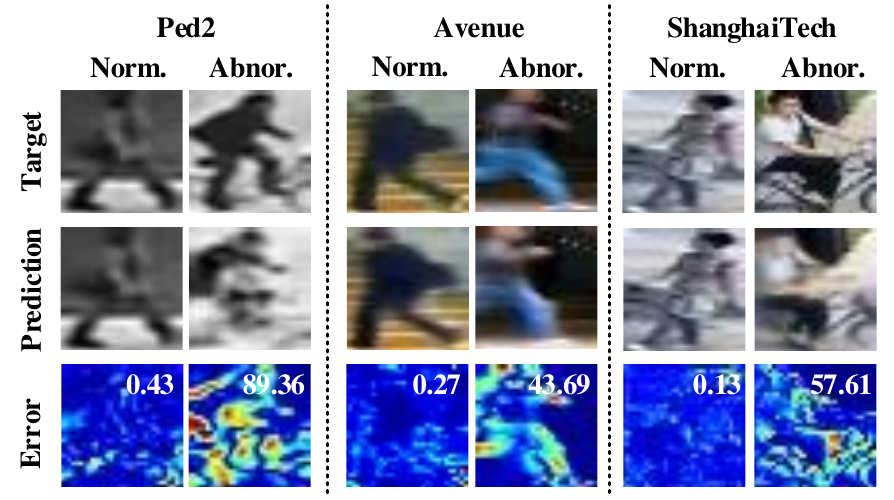}
	\caption{Visualization examples of our model, including the ground truth frames (Target), completed frames (Prediction), and completion errors (Error).}
	\label{p6}
\end{figure}

\subsection{Ablation Studies}
We perform corresponding ablation studies to analyze the impact of different components of our framework on the UCSD Ped2 dataset, including joint correspondence modeling, semantics alignment loss, and MGSM. The results are shown in Table \ref{table:ablation}. We can see that joint correspondence modeling (Index 3) obtains an improvement compared to single appearance normality (Index 1) and single motion normality (Index 2), respectively, indicating the effectiveness of the correlation between appearance and motion for VAD. Additionally, the proposed semantics alignment loss (Index 4) can explicitly strengthen the correlation, which helps to get a significant AUC improvement. Index 5-6 explore the effectiveness of MGSM, and the result shows that the AUC performance is significantly enhanced because MGSM realizes a tradeoff between the good reconstruction of normal data and the poor reconstruction of abnormal data vital for VAD.

To evaluate the adaptability of MGSM to the complex scene, we select the challenging ShanghaiTech dataset to study the robustness of the number of memory items. We conduct the experiments using different memory size settings and show the results in Table \ref{table:items}, noting that all three MGSMs are of the same size. Notably, as the memory size is tuned, AUC changes steadily. A small or large memory pool further does not lead to obvious performance degradation, indicating that the proposed MGSM has the benefit of good reconstruction on normal samples without worrying about the curse of `overgeneralizing', which would not be possible by simply revising the model size. For more discussion and supplementary experiments about MGSM, please see the \textbf{appendix}.

\begin{table}
\footnotesize
\centering
\caption{
Ablation studies of each component in our framework on the UCSD Ped2 dataset.
}
\label{table:ablation}
\begin{threeparttable}
    \begin{tabular}{c|ccccc|c}
    \hline
             Index   & $E_{\varphi}$ & $E_{\theta}$ & FFRP & $L_{align}$ & MGSM & AUC (\%)\\ \hline
      1            & \CheckmarkBold     &        &    &  &   &  91.9 \\
     2            &       & \CheckmarkBold            &   &    &   &  87.8        \\
     3   &  \CheckmarkBold  & \CheckmarkBold  & \CheckmarkBold &   &   &  93.8 \\
     4   &  \CheckmarkBold  & \CheckmarkBold  & \CheckmarkBold & \CheckmarkBold  &   &  98.2 \\
     5            & \CheckmarkBold     &        &    &  &  \CheckmarkBold &  95.6 \\
     6   &  \CheckmarkBold  & \CheckmarkBold  & \CheckmarkBold & \CheckmarkBold  & \CheckmarkBold  & 99.6  \\
    \hline
    \end{tabular}
    \begin{tablenotes}
			\scriptsize
			\item FFRP = flow feature replacement for prediction.
			\item MGSM = multi-scale memory-guided suppression module.
		\end{tablenotes}
\end{threeparttable}
\end{table}

\begin{table}
\footnotesize
\begin{center}
\caption{
AUC (\%) analysis on the quantity of memory items in MGSM on the ShanghaiTech dataset.
}
\label{table:items}
\begin{tabular}{c|cccccc}
\hline
         Number   &  20 & 40  & 80  & 160 & 200 & 240\\ \hline
 AUC   &  75.7 & 75.4  & 75.9  & 76.2 & 75.8 & 75.7\\ 
\hline
\end{tabular}
\end{center}
\end{table}

\section{Conclusion}
In this paper, we propose a multi-level memory-augmented appearance-motion correspondence framework for video anomaly detection. The correlation between appearance and motion normality is explicitly modeled via the proposed appearance-motion semantics alignment loss and semantics replacement for prediction. Meanwhile, the proposed MGSM not only achieves the tradeoff between the good reconstruction of normal data and poor reconstruction of abnormal data but also overcomes the problem of performance degradation in complex scenes. Extensive experimental results on three benchmark datasets demonstrate that our method performs better than state-of-the-art approaches.

\bibliographystyle{IEEEbib}
\bibliography{icme2023template}

\newpage
\section{AppendiX}

\begin{figure*}[!htbp]
\centering
\includegraphics[width=170mm]{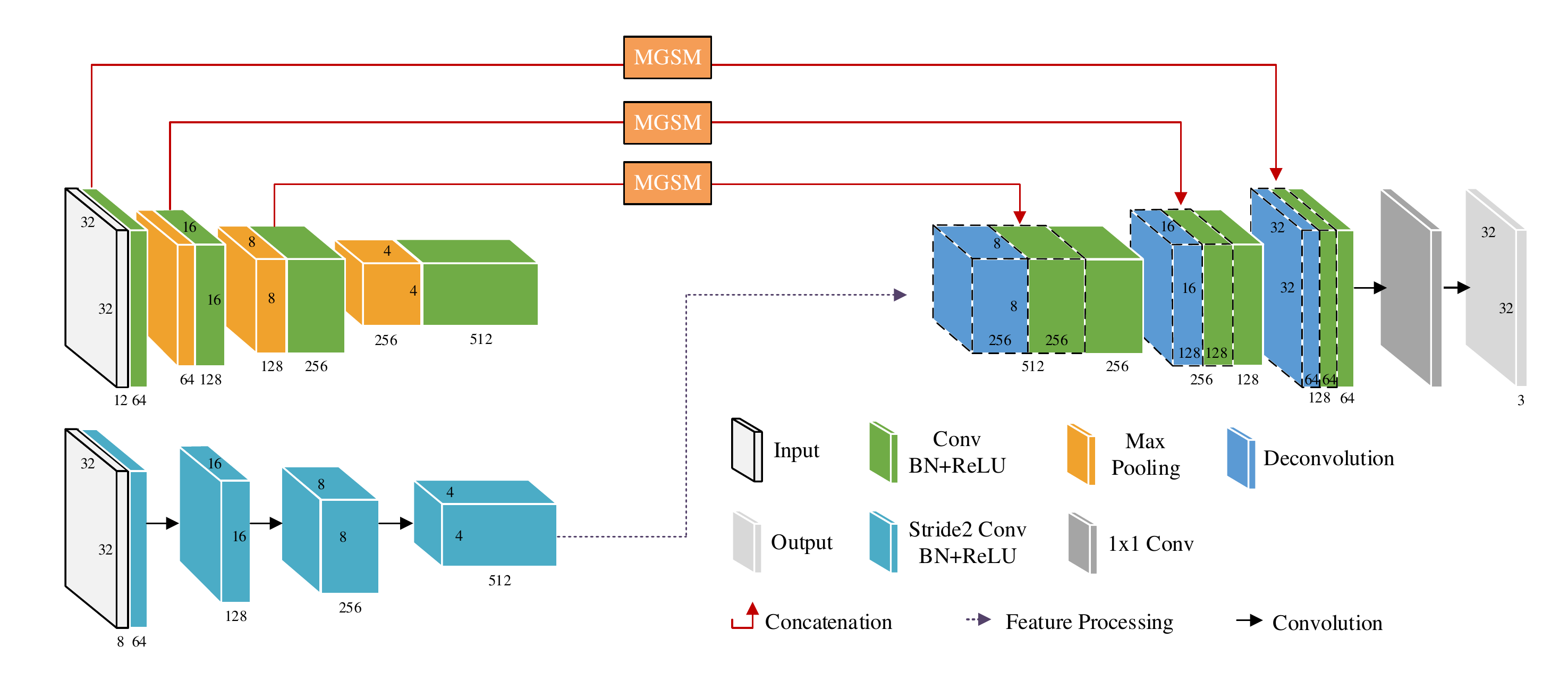}
\caption{Detailed network architecture of the proposed framework.
}
\label{fig:model}
\end{figure*}
\subsection{Detailed Network Design}
In Figure 4, we illustrate the detailed network architecture of the proposed framework. Each cube in the network is the output feature maps for the corresponding layer. As shown, we have two encoders and one decoder. The kernel size of all convolutional layers in the network is fixed to 3×3. Similar to \cite{liu2018future}, the appearance encoding layer is composed of a maxpooling layer, a convolution layer, a batch-normalization layer and a ReLU layer sequentially. The motion encoding layer consists of a stride-2 convolution layer, a batch-normalization layer and a ReLU layer sequentially. The up-sampling layer is implemented by stride-2 transposed convolution. Every down-sampling encoded feature is processed by the proposed MGSM and then sent to the decoder by skip-connection for prediction.
Our model contains 4 levels in total, and the corresponding feature map sizes of each level are (32,32,64), (16,16,128), (8,8,256) and (4,4,512), respectively. please see the code for more details.

\begin{figure*}[!htbp]
\centering
\includegraphics[width=170mm]{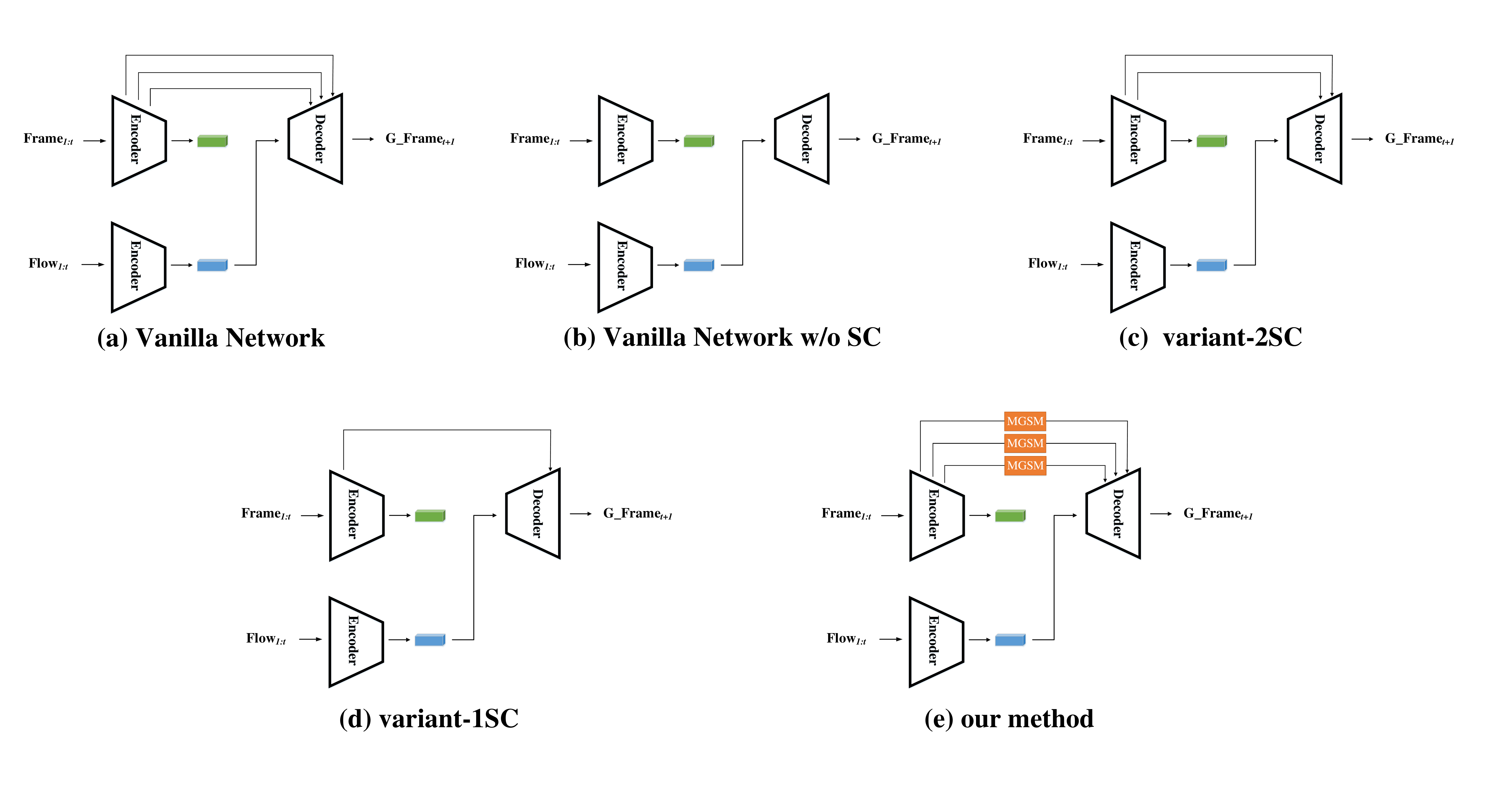}
\caption{Some variants about skip-connection in VAD.
}
\label{fig:dicussion}
\end{figure*}

\subsection{Discussion about Skip-Connection in VAD}
In this subsection, we conduct additional experiments about more variants of our method to analyze the impact of skip-connection. The architecture of some variants is illustrated in Figure \ref{fig:dicussion}, and the results are shown in Table \ref{t2}. As we can see, compared with Vanilla Network, which is equipped with fully skip-connection, Vanilla Network w/o SC is in the dilemma of severe performance degradation. It means that skip-connection plays an important role in improving the reconstruction ability of the model, which meets the need for prediction-based VAD methods. At the same time, we argue that the abuse of skip-connection is prone to learning an identity mapping, which leads to the good reconstruction of abnormal data. For the purpose of directly weakening the impact of the skip-connection, we decide to remove some connections. Meanwhile, to avoid excessive inhibition of its effect, we reserve the outermost skip connection. So variant-2SC and variant-1SC are designed. Surprisingly, when the number of removed connections increases, the performance of the model on the Ped2 and ShanghaiTech datasets improves, but the performance on the Avenue dataset decreases. We think that different types of scenarios have different requirements for reconstruction capabilities. Compared with Ped2 and ShanghaiTech datasets, the scale of the foreground region on the Avenue dataset is very large, so the requirements for reconstruction ability on the Avenue dataset are higher. It demonstrates that the unsuitable reconstruction capacity for different scenes caused by skip-connection, is the major limitation of VAD. In order to improve the adaptability to different complex scenes, we fix the MGSM module into the skip connection to adaptively balance the reconstruction of the normality and anomaly.

\begin{table}
\footnotesize
    \begin{center}
	\caption{AUROC (\%) comparison between our method and different variants.}
	\label{t2}
	\begin{tabular}{c|ccc}
    \hline
    Model            &   UCSD Ped2 & CUHK Avenue & ShanghaiTech \\ \hline
			 Vanilla Network & 93.8          & 92.6            & 74.3           \\
                Vanilla Network w/o SC  & 81.6          & 73.2          & 57.5          \\
			 variant-2SC       &  94.9         &  91.3         &   74.6        \\
                variant-1SC    & 96.8          & 89.4             & 75.2      \\
    \textbf{Our method}            & \textbf{99.6}      & \textbf{93.8}       & \textbf{76.3}        \\ \hline
	\end{tabular}
	\end{center}
\end{table}

\end{document}